# Deep Learning-based Event Data Coding: A Joint Spatiotemporal and Polarity Solution

Abdelrahman Seleem, *Graduate Student Member, IEEE*, André F. R. Guarda, *Member, IEEE*, Nuno M. M. Rodrigues, *Senior Member, IEEE*, Fernando Pereira, *Fellow, IEEE*

**Abstract** — Neuromorphic vision sensors, commonly referred to as event cameras, generate a massive number of pixel-level events, composed by spatiotemporal and polarity information, thus demanding highly efficient coding solutions. Existing solutions focus on lossless coding of event data, assuming that no distortion is acceptable for the target use cases, mostly including computer vision tasks such as classification and recognition. One promising coding approach exploits the similarity between event data and point clouds, both being sets of 3D points, thus allowing to use current point cloud coding solutions to code event data, typically adopting a two-point clouds representation, one for each event polarity. This paper proposes a novel lossy Deep Learning-based Joint Event data Coding (DL-JEC) solution, which adopts for the first time a single-point cloud representation, where the event polarity plays the role of a point cloud attribute, thus enabling to exploit the correlation between the geometry/spatiotemporal and polarity event information. Moreover, this paper also proposes novel adaptive voxel binarization strategies which may be used in DL-JEC, optimized for either quality-oriented or computer vision task-oriented purposes which allow to maximize the performance for the task at hand. DL-JEC can achieve significant compression performance gains when compared with relevant conventional and DL-based state-of-the-art event data coding solutions, notably the MPEG G-PCC and JPEG Pleno PCC standards. Furthermore, it is shown that it is possible to use lossy event data coding, with significantly reduced rate regarding lossless coding, without compromising the target computer vision task performance, notably event classification, thus changing the current event data coding paradigm.

## I. INTRODUCTION

IN the golden age of multimedia, event cameras are emerging as revolutionary devices for capturing fast-moving scenes with high precision and efficiency. Unlike traditional frame-based cameras that capture two-dimensional (2D) images, event cameras asynchronously record changes (positive or negative) in individual pixels intensities, generating data commonly called *events*, enabling high-speed, high dynamic range and low-latency data acquisition [1]. This unique capability and associated characteristics make them ideal for applications such as robotics, autonomous vehicles [2], and augmented reality [3], where real-time responsiveness and efficiency are critical.

Event cameras register events which are independently generated for each pixel position in the sensor. Each event in the event sequence is represented by a four-dimensional (4D) tuple $(x, y, t, p)$, where $(x, y)$ are the spatial coordinates of each triggering pixel position, $t$ is the timestamp of when the event occurs, and $p$ is the polarity, indicating whether the brightness for that specific sensor pixel increased or decreased. This new way of capturing visual data provides very detailed information about the scene, notably with very high temporal resolution and dynamic range, what is critical for some specific application scenarios [4]. However, the high temporal and spatial resolutions produce sequences with a massive number of events, which poses challenges for storage and transmission, thus demanding efficient coding solutions in order practical implementations are viable.

Currently, event data coding solutions mostly adopt conventional lossless coding methods, targeting the use of the original event data in the computer vision tasks, without accepting any data degradation. As for other modalities, lossless coding is commonly needed when it is essential to preserve every bit of the original data without any loss, such as for medical imaging and other critical data; naturally, this comes at the cost of much increased rate and should only be used if quality controlled lossy coding is unacceptable. Lossy coding accepts some degradation in the original information.

This work was supported by the Fundação para a Ciência e a Tecnologia (FCT, Portugal) through the research project PTDC/EEI-COM/1125/2021, entitled "Deep Learning-based Point Cloud Representation", and by FCT/MECI through national funds and when applicable co-funded EU funds under UID/50008: Instituto de Telecomunicações.

A. Seleem is with Instituto Superior Técnico - Universidade de Lisboa, Lisbon, Portugal; Instituto de Telecomunicações, Portugal; and South Valley University, Qena, Egypt (e-mail: a.seleem@lx.it.pt).

A. F. R. Guarda is with Instituto de Telecomunicações, Portugal (e-mail: andre.guarda@lx.it.pt).

N. M. M. Rodrigues is with ESTG, Politécnico de Leiria, Leiria, Portugal; and Instituto de Telecomunicações, Portugal (e-mail: nuno.rodrigues@co.it.pt).

F. Pereira is with Instituto Superior Técnico - Universidade de Lisboa, Lisbon, Portugal; and Instituto de Telecomunicações, Portugal (e-mail: fp@lx.it.pt).

For modalities consumed by humans, such as images, video and point clouds (PCs) the degradation has often limited or even negligible impact, notably due to the limitations of the human visual system. Until recently, event data has been only losslessly coded, considering that no degradation could be acceptable, notably because event data is typically consumed by computer vision tasks and not directly by the human visual system; however, recent works, and notably this paper, demonstrate that lossy coding may be acceptable if not significantly impacting the computer vision tasks performance, while benefiting from significant rate reductions. Nevertheless, this requires the encoder to control the rate-quality trade-off, limiting signal degradation to an acceptable level for the task at hand.

The need for efficient event data coding solutions has been recognized by the JPEG (Joint Photographic Experts Group) standardization group which has launched the JPEG XE project targeting *"the creation and development of a standard to represent Events in an efficient way allowing interoperability between sensing, storage, and processing, targeting machine vision and other relevant applications.*" [5]. In July 2024, a Call for Proposals has been issued to "*receive input contributions for technology to losslessly represent event data as an interchangeable format.*" [6]. While lossless coding is the target of the first phase of JPEG XE, lossy coding is already included in the requirements and will be considered in a second phase [5].

One promising approach to efficiently code event data involves the representation of event data as one or more PCs. A PC consists of a set of points in the three-dimensional (3D) space defined by their coordinates (*x, y, z*), referred to as the PC *geometry*. Besides geometry, a PC may include additional information about the 3D points, the *attributes*, such as color and normal vectors. Due to the obvious similarity between event data and PCs, the event spatial and temporal components can be represented as the PC geometry by perceiving the voxel coordinates as (*x, y, z=t*), where the timestamp component is mapped to the z-axis. Since this geometry-only PC representation does not include the event polarity information, some event coding solutions have been designed to code the event data as two separate geometry-only PCs, one for each polarity [7]-[12].

Recognizing the critical need for efficient Point Cloud Coding (PCC) in application scenarios where interoperability is essential, the JPEG standardization group has recently concluded the specification of the JPEG Pleno Learning-based Point Cloud Coding standard, known as JPEG Pleno PCC (JPEG PCC) [13][14][15]. This first Deep Learning (DL)-based PCC standard outperforms the available conventional PCC standards specified by the Moving Picture Experts Group (MPEG), notably geometry-based PCC (G-PCC) for coding static PCs and video-based PCC (V-PCC) for coding dynamic PCs [16], particularly for the geometry of dense static PCs [13][14][15].

While component scalability is an important requirement for MPEG and JPEG PCC standards [14][15], where decoding PC geometry without decoding the associated attributes is useful for many use cases (e.g., PC classification), the same does not happen for event data, since processing spatiotemporal event information without polarity is not required by any identified use case. This is confirmed by the absence of a component scalability requirement in the JPEG XE requirements [5] contrary to what happened for the JPEG PCC requirements [17]. This stimulates and enables the development of joint coding solutions for event data, where spatiotemporal and polarity information are coded together instead of in a component scalable manner, i.e., spatiotemporal information independent from the polarity information, with the potential for increased compression efficiency by exploiting the correlation between all event data components.

In this context, this paper proposes a novel lossy Deep Learning-based Joint Event data Coding (DL-JEC) solution where the event data is represented as a single PC, with (*x, y, z=t*) corresponding to the PC geometry, and the polarity *p* corresponding to a PC attribute. To the best of the authors' knowledge, this is the first DL-based joint event data coding solution in the literature using a single PC for coding the full information associated to an event data sequence/object. The proposed DL-JEC solution adopts a PC representation for the event data and is designed as an extension to the lossy JPEG PCC standard [13][14][15], improving the rate-distortion (RD) performance when compared with state-of-the-art PC-based event data codecs.

While event data is commonly used for machine-oriented consumption, such as classification, its application for man-oriented consumption is not common; this is different from PC coding where human visualization is most important. This paper shows that by carefully tailoring the joint coding model and the decoder binarization process, which suitably controls the number of reconstructed points/events, the proposed DL-JEC solution can be optimized for coding performance (targeting high reconstruction fidelity) or for improved performance of a given application (e.g., a computer vision task); in this paper, the event data classification task and the associated performance are adopted to demonstrate these claims.

In this context, the novel technical contributions of this paper are:

- A novel lossy event data coding solution, DL-JEC, extending the JPEG PCC standard to jointly code the spatiotemporal and polarity information for event data represented as a single PC, instead of two PCs, one per polarity. This is the first DL-based joint coding solution proposed for event data.
- Novel adaptive voxel binarization strategies for the reconstruction of the event data which are tailored to maximize the performance, either in terms of compression or a target computer vision task, e.g., classification. This makes the codec rather flexible in terms of target application scenarios.
- A novel quality metric for assessing (lossy) reconstructed event data, named Peak Signal-to-Noise Ratio Event-to-Event Distance (PSNR E2E), able to measure the impact of compression artifacts.

The novel technical contributions have been assessed under solid and representative experimental conditions, notably:

- Experimental assessment of the DL-JEC compression performance, showing significant gains when compared to conventional and DL-based PCC methods, notably MPEG G-PCC (lossy and lossless) and JPEG PCC (lossy).
- Experimental assessment of the DL-JEC classification performance and relevant benchmarks for reconstructed event data using the state-of-the-art DL-based Event Spike Tensor (EST) [18] classifier. Using the novel binarization strategies, the proposed DL-JEC solution shows substantial performance gains over the adopted benchmarks, for the widely adopted Neuromorphic-Caltech101 (N-Caltech101) [19] event dataset.
- Experimental assessment of DL-JEC generalization power by evaluating the compression and gesture recognition performances on a different test dataset, notably the International Business Machines Dynamic Vision Sensor Gesture (IBM DVS Gesture) [20] event dataset. The results show that DL-JEC offers good generalization capabilities, still showing competitive performance regarding the selected benchmarks.

A repository with the relevant reconstructed event sequences from the two adopted test datasets and the full experimental results will be made publicly available to enable the research community to use DL-JEC codec as benchmark for new event data coding solutions.

This paper is organized as follows: Section II offers a brief review of lossy event data coding solutions; Section III describes the relevant background technologies for event data coding and classification; Section IV presents the adopted DL-based event data coding architecture; Section V details the proposed DL-JEC solution, including the coding model and binarization optimizations strategies; Section VI describes the experimental setup; Section VII reports and discusses the performance assessment; finally, Section VIII reports the conclusions and future work.

## II. REVIEW OF LOSSY EVENT CODING SOLUTIONS

This section presents a brief review of lossy event data coding – the key topic of this paper. The review is organized according to the data structure used to represent the event sequences, notably sequence-based, frame-based, and 3D point set-based, corresponding to one-dimensional (1D), 2D, and 3D representations, respectively. Particular attention is given to event data coding solutions adopting a PC-based representation, as this is the main focus of this paper.

The sequence-based coding solutions process the raw input data as a 1D array, where each array element represents one event from the raw input event sequence. These elements are ordered based on one or more criterion, usually the timestamp value associated to each array element. Such 1D arrays are suitable for lossless coding, by preserving all the information about the events' components, or lossy coding, by aggregating or sampling/discarding data to improve the compression performance [21].

An example of a sequence-based lossy coding solution is the Flow-Based lossy Compression (FBC) method proposed in [22], which is designed for real-time lossy coding of event data acquired with Dynamic and Active Pixel Vision Sensors (DAVIS) cameras. FBC works in two modes which are selected depending on the quality of a prediction made using a short-term optical flow. A lightweight encoder computes the optical flow and confidence scores to decide whether to use the raw mode, which encodes the raw events directly, or the predictive mode, which encodes the events using flow-driven predictions without residue. The decoder reconstructs the events using either the raw data or the flow prediction information, thus reducing the coding rate without significantly compromising quality. FBC was tested on five real-world sequences (from the DAVIS 240C dataset [23], [24] and the Multi Vehicle Stereo Event Camera (MVSEC) dataset [25]) over 30 ms, achieving an average Compression Ratio (CR) of 2.81, while maintaining minimal errors (approximately 0.48 ms in time and 3.07 average distance). When FBC is combined with Lempel-Ziv-Markov chain Algorithm (LZMA) lossless coding, the reported CR values increase across different time periods used for the predictive mode (30, 60 and 90 ms), with average CR values ranging from 10.45 to 17.24. Another example of sequence-based coding solution is the Event Vision Stream Compression (EVSC) framework [26], which proposes a modular structure with three main stages: irrelevancy reduction through lossy filtering, spatiotemporal event prediction, and lossless entropy coding. In the first stage, optional lossy filters, such as random-drop filters, remove events that are unlikely to help with downstream tasks. In the second stage, the system reduces redundancy by predicting future events from past spatio-temporal patterns using Same-Event Prediction, and then computing residuals between the predicted and actual events. In the final stage, these residuals are treated as low-entropy symbols and compressed with standard entropy coders such as Huffman or arithmetic coding. The effect of event filtering on a semantic segmentation task was evaluated on the D-Stereo Event Camera Dataset for Driving Scenarios (DSEC) [27] semantic dataset using the mean intersection-over-union (mIoU) metric. The results show that accuracy and mIoU remain nearly unchanged at 500 Mbps and only slightly decrease at 200 Mbps, while more noticeable drops occur at 100 Mbps, indicating that task-specific filtering strategies can help limit these performance losses [26].

Event data coding solutions using a frame-based data structure convert the event sequences into sets of 2D frames where each 2D frame has the sensor's spatial resolution and includes per-pixel temporally aggregated values. Different temporal aggregation methods may be used, including pixel-wise accumulation of polarity, counting the number of events at each pixel over a certain time interval (event counting), or sampling elements over time (temporal sampling) [21][28][29]. As a result, even if the frames are losslessly coded, the overall coding solution is lossy due to the temporal aggregation. Organizing event data into 2D frames enables the use of standard (and exhaustively optimized) video coding solutions/standards, which allow a very efficient exploitation of the spatial and temporal correlations to improve the compression performance [21].

An example of a frame-based coding solution is the low-precision sparse autoencoder (AE) method proposed in [29], where raw event sequences are converted into binary frames which are then processed by an AE architecture based on sparse convolutional layers; the specific details of the conversion process from event data to binary frames are not explicitly described [29]. The AE uses quantized convolutional operations at 2-bit and 4-bit levels during training to reduce the computational complexity. The reconstructed images are then used for downstream tasks such as event-based classification and object detection. Experimental results based on the IBM DVS Gesture dataset [20], a neuromorphic version of the classic MNIST (MNIST-DVS) [19], and Gen1 Neuromorphic Cars (N-CARS) [30] datasets show that this coding solution achieves CR values of 12.1, 29.1 and 18.1, with small accuracy drops for the classification task of 2.0%, 3.0% and 3.8%, respectively. For the object detection task, experimental results on the Gen1 Automotive Detection (Prophesee-Gen1) dataset [31] show an average CR value of 11.9 with a drop of 0.07 in mean average precision (mAP), compared to the uncompressed baseline.

Another frame-based coding solution [32] aggregates event data information over fixed time intervals into two polarity-based frames, representing Positive (POS) and Negative (NEG) event counts per pixel, effectively forming spatial histograms. These histograms are concatenated horizontally into *superframes*. A sequence of such superframes generated over time constitutes a 3D representation of the event sequence. This 3D structure is encoded using a Deep Belief Network (DBN) to extract low-dimensional latent features, which are then compressed using entropy-based Huffman coding. Experimental results on three event sequences from the DAVIS 240C dataset [23] show that the CR gains vary with the temporal aggregation period.

Event data coding solutions using a 3D point set-based data structure organize the event data as a set of points in 3D space, where each point is defined by the spatial coordinates $(x, y)$ and a timestamp $t$, treating $t$ as the third spatial dimension to create a 3D geometry. Additional information, such as polarity, may also be included as attributes to the geometry. This data structure is typically sparse and may preserve (or not) the original event information, by avoiding (or not) temporal aggregation and subsampling. The 3D point set-based data structure is well-suited for coding using PCC methods, notably the G-PCC [16] and JPEG PCC [15] standards, which are well-suited for handling the sparsity and correlation of 3D event data points represented as a PC. Existing solutions often code event data as two separate PCs (one for each polarity) to facilitate processing and directly use the available PCC solutions. These solutions may adopt either lossless [7][8][11][12] or lossy [9][10][33] coding approaches, using both conventional [9][33] or DL-based models [10].

In [9], the event sequences are encoded using the conventional G-PCC Octree in lossy coding mode, and assessed on various computer vision tasks, such as object recognition, optical flow, and depth estimation. Results for the N-Caltech101 [19], MVSEC [25] and DAVIS 240C [23] datasets show that high CR values can be achieved while maintaining high performance for the computer vision tasks. In [33], a modified G-PCC encoder is proposed for lossy compression of event data by representing accumulated events as single PC. Events are aggregated over fixed time intervals into polarity-separated 2D histograms, which are converted into a multivariate stream containing spatial coordinates, time index (including the time aggregation interval), event count, and polarity. The single PC is formed by treating the spatial coordinates and the time index as PC geometry, while the event count and polarity are used as PC attributes, thereby enabling the joint coding of the event data components. Compression performance is evaluated using spatial and temporal distortion metrics, including 2D PSNR and timing error. Experiments on the DAVIS 240C dataset [23] show that this method achieves high CRs with low distortion, maintaining 2D PSNR values above 42 dB even at very low bitrates [33]. Unlike the DL-JEC solution proposed in this paper, this conventional coding approach only addresses compression purposes and cannot support compressed domain computer vision tasks.

In [10], a DL-based architecture for efficient lossy event data coding is proposed, using JPEG PCC to code the two generated PCs (one per polarity). This architecture, which is an inspiration for this paper, uses an Event to Two PCs Conversion module at the encoder and a Two PCs to Event Conversion module at the decoder, which may adopt a varying Temporal Scaling Factor (TSF) to convert an event data sequence into two PCs (one per polarity). This coding solution was evaluated on the N-Caltech101 [19] and CIFAR10-DVS [34] datasets, with the results demonstrating that JPEG PCC can significantly reduce the coding rate compared to G-PCC lossless, while maintaining a classification performance similar to the one associated to the original event data.

In summary, despite the recent progress made in event data coding using different data structures, the compression performance still needs to be improved, notably using DL-based technologies. Moreover, it remains to demonstrate that event data lossy coding solutions may be used for computer vision tasks, without impactful performance penalty, while benefiting from significant rate reductions for diverse application scenarios with different latency, random access and compression requirements.

## III. BACKGROUND TECHNOLOGIES FOR EVENT DATA CODING AND CLASSIFICATION

This section introduces the most relevant background technologies for the event data coding and classification solution proposed in this paper, notably JPEG PCC coding and EST event data classification [10].

### A. JPEG PCC Overview

This subsection briefly presents the DL-based JPEG PCC standard codec, which includes the foundational coding model for the proposed lossy DL-JEC solution, proposed in detail in sections IV and V. JPEG PCC is also one of the codecs used in the two PCs event coding data solution proposed in [10]. For more details on the JPEG PCC standard, please refer to [13][14][15].

The JPEG PCC high-level architecture includes a first module to divide the input voxelized PC into 3D blocks, enabling random access and reducing the required computational resources. Optionally, the PC blocks can be down-sampled, which enhances compression efficiency for sparse PCs or facilitates achieving lower rates.

The JPEG PCC core component is the DL coding model, designed to code 3D blocks of binary voxels from the input voxelized PC, where each voxel is either occupied (1) or empty (0). The JPEG PCC coding model includes an AE, augmented by a variational hyperprior model. The AE compresses the input PC block into a compact latent representation, which is then quantized and entropy-coded using a dynamically estimated probability distribution, characterized by the mean and standard deviation values, provided by the variational hyperprior model.

At the decoder, the latent representation is processed through the AE decoding layers to reconstruct the PC block. This process outputs voxel occupancy probabilities, which are binarized using a threshold optimized at the encoder according to some relevant criterion, e.g., quality metric, thus determining the number of reconstructed points. The DL coding model is trained end-to-end for various target rates under the JPEG PCC Common Training and Test Conditions (CTTC) [35].

Following decoding, an up-sampling step reverses the prior down-sampling, if applied. This up-sampling process may include a DL-based super-resolution model to enhance the quality of the reconstructed PC block. In terms of RD performance, JPEG PCC surpasses both G-PCC Octree and V-PCC Intra [16] for geometry coding of dense static PCs [13][14][15].

### B. EST Classifier Overview

This subsection briefly presents the adopted DL-based event data classifier, specifically the Event Spike Tensor (EST) classifier [18], which is considered one of the state-of-the-art solutions for event-based data classification.

The EST classifier adopts an event representation compatible with Convolutional Neural Network (CNN) architectures, which is learned end-to-end alongside the classification task, thus optimizing the representation. The input event data is converted into a 4D EST representation, using 9 bins for the temporal dimension and concatenating polarity information, resulting in an image-like format of dimensions 18×224×224, where the 18 channels correspond to single component 224×224 resolution images [18].

The EST classifier architecture is adapted from ResNet-34 [36], with modifications to the first and last layers. The EST classifier achieves performance improvements of approximately 12% for event data classification compared to state-of-the-art methods, including event frames [37], voxel grids [38], and histograms of averaged time surfaces (HATS) [30].

Since the pre-trained EST classifier model was not publicly available, the EST classifier model was trained in the context of this paper using the training and validation splits of the N-Caltech101 [19] and IBM DVS Gesture [20] datasets. The training used a total of 30 epochs and the Adam optimizer with an initial learning rate of $10^{-4}$, which was halved every 10,000 iterations. The training used cross-entropy as the loss function to measure the discrepancy between predicted and ground truth classification labels.

## IV. Proposed Lossy DL-based Joint Event Data Coding: DL-JEC Architecture

While coding event data as two PCs leverages the relationship between events of the same polarity, it does not take advantage of the relationship between events of different polarities. Coding event data as a single PC addresses this limitation by including both polarities in a unified representation.

DL-JEC is the first DL-based coding solution in the literature to jointly code spatiotemporal and polarity information within a single PC representation. DL-JEC is a pioneering approach proposed in opposition to the two PCs approach [9][10], which has been used in previous works. By considering the specific characteristics of event data consumption, notably the fact that event data containing only spatiotemporal information, without polarity, is not commonly useful, DL-JEC does not adopt a component scalable approach, in which some components are coded independently from others, as it happens for the MPEG and JPEG PCC standards [13][15] where geometry is first independently coded from texture which is after coded in a component scalable way.

This section presents the novel DL-JEC high level architecture for lossy coding of event data represented as a single PC. The key novel components for the proposed DL-JEC solution are presented in detail in the next section.

The architecture for coding event data as a single PC, as adopted by DL-JEC, is depicted in Fig. 1b while Fig. 1a shows the previous two PCs coding architecture. The DL-JEC architecture includes the following modules:

1. *Event to Single PC Conversion* – Converts the input event sequence (*x, y, t, p*) into a single PC. The main tools involved in the Event to Single PC Conversion are:
   - *Temporal Scaling* – Scales the timestamps of the event sequence/object using a selected TSF value, while the spatial coordinates remain unchanged. Temporal coordinates are scaled as $z = t \times TSF$, where $t$ is the event timestamp; this module reduces the precision of the event data temporal component depending on the selected TSF value.

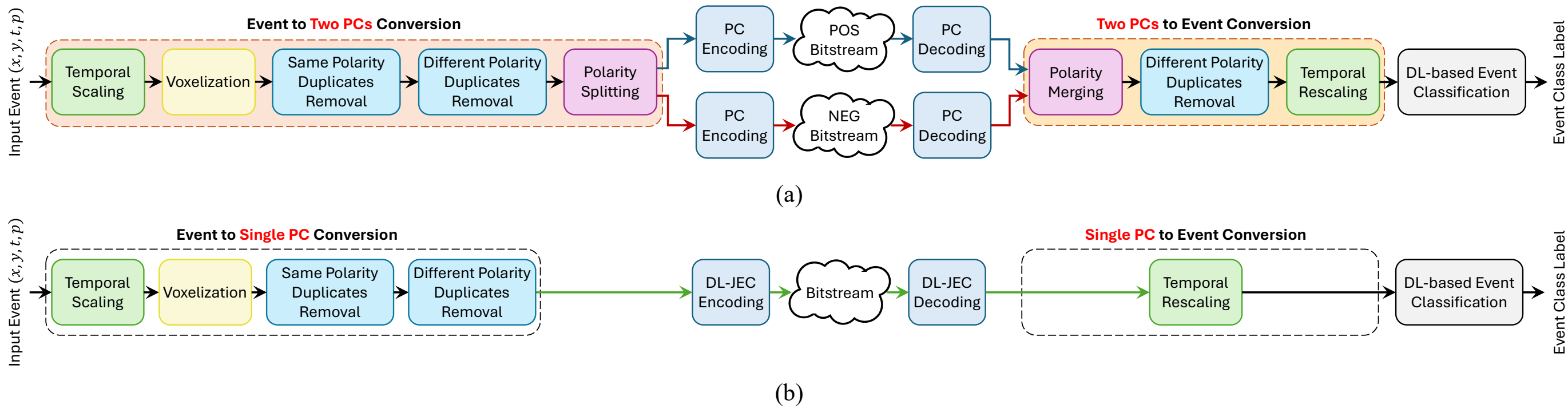

Fig. 1. DL-based event data coding and classification architecture: (a) as two geometry-only PCs; (b) as a single geometry+attribute PC as adopted in DL-JEC.

- *Voxelization* – Converts the scaled temporal coordinates into integers as required for voxel-based PCC solutions (like G-PCC, JPEG PCC and DL-JEC); spatial coordinates are not affected as they are already integers.
- *Same Polarity Duplicates Removal* – Removes duplicate events at same spatiotemporal position and polarity, which may result from the lossy voxelization step.
- *Different Polarity Duplicates Removal* – Removes duplicate events at the same spatiotemporal position with different polarities, which may result from the lossy voxelization step, thus ensuring that each single point/event exists for one single polarity. In such cases, only the event with the same polarity of its Nearest Neighbor (NN) is kept.

2. *DL-JEC Encoding* – The single PC resulting from the Event to Single PC Conversion module is encoded using the novel DL-JEC coding solution, detailed in the next section.

The architecture for decoding event data from a single PC proceeds as follows:

3. *DL-JEC Decoding* – The event data is reconstructed from the bitstream using the decoding part of the DL-JEC model, described in detail in the next section.
4. *Single PC to Event Conversion* – Converts the single reconstructed PC into a reconstructed event sequence/object. The only tool in this conversion is the *Temporal Rescaling* which restores the temporal event information from the reconstructed $z$ coordinates, $z'$, as $t' = z'/TSF$ where TSF is the same value as used for the Temporal Scaling.

The DL-JEC solution proposed in this paper may be compared with a two PCs coding solution, which uses the architecture represented in Fig. 1a. The first key difference appears at the event data encoder side. The Event to Two PCs Conversion module shares most of the tools used in the Event to Single PC Conversion module, but uses a different final module, the Polarity Splitting tool which is used to convert the voxelized PC data into two PCs based on the polarity. The duplicates removal process is the same for the conversion modules in the two architectures, as duplicated points/events with different polarities are not considered useful information.

The second key difference is at the core coding module. The standard PC Encoding and PC Decoding tools (e.g., G-PCC lossless, G-PCC Octree or JPEG PCC) used in the previous solution to encode and decode the event data, represented as two geometry-only PCs, are replaced by the proposed DL-JEC Encoding module, detailed in the next section, which jointly encodes the spatiotemporal and polarity information, combined into a single PC data structure.

The third key difference arises at the PC to Event Conversion level, performed at the decoder. The two reconstructed PCs are converted back into event data by using a Two PCs to Event Conversion module. This module includes a Polarity Merging tool, to merge the two reconstructed PCs and add the associated polarity information back to the reconstructed events, and a Different Polarity Duplicates Removal tool to ensure that each point/event has a single polarity, since the two independent PCs decoding processes may create co-located events with different polarities. This is performed by defining the duplicate point polarity based on the (highest) voxel occupancy probabilities for both polarities, estimated by the JPEG PCC decoding model, or using a NN process for G-PCC Octree lossy reconstructed event data.

The DL-based Event Classification module is common to both coding pipelines and used in this paper to compare the performance of the various decompressed domain event data classification pipelines, notably using event data represented as one or two PCs.

## V. DL-JEC Coding Model and Binarization Optimization Strategies

This section details the proposed DL-JEC coding model and associated binarization optimization strategies, represented in Fig. 2. Although the novel DL-JEC solution and the previous JPEG PCC standard share a similar high-level coding architecture, DL-JEC and JPEG PCC are fundamentally distinct codecs. Innovations arise in several areas, notably the DL coding model, loss function, and binarization optimization, which are presented in detail in the following subsections.

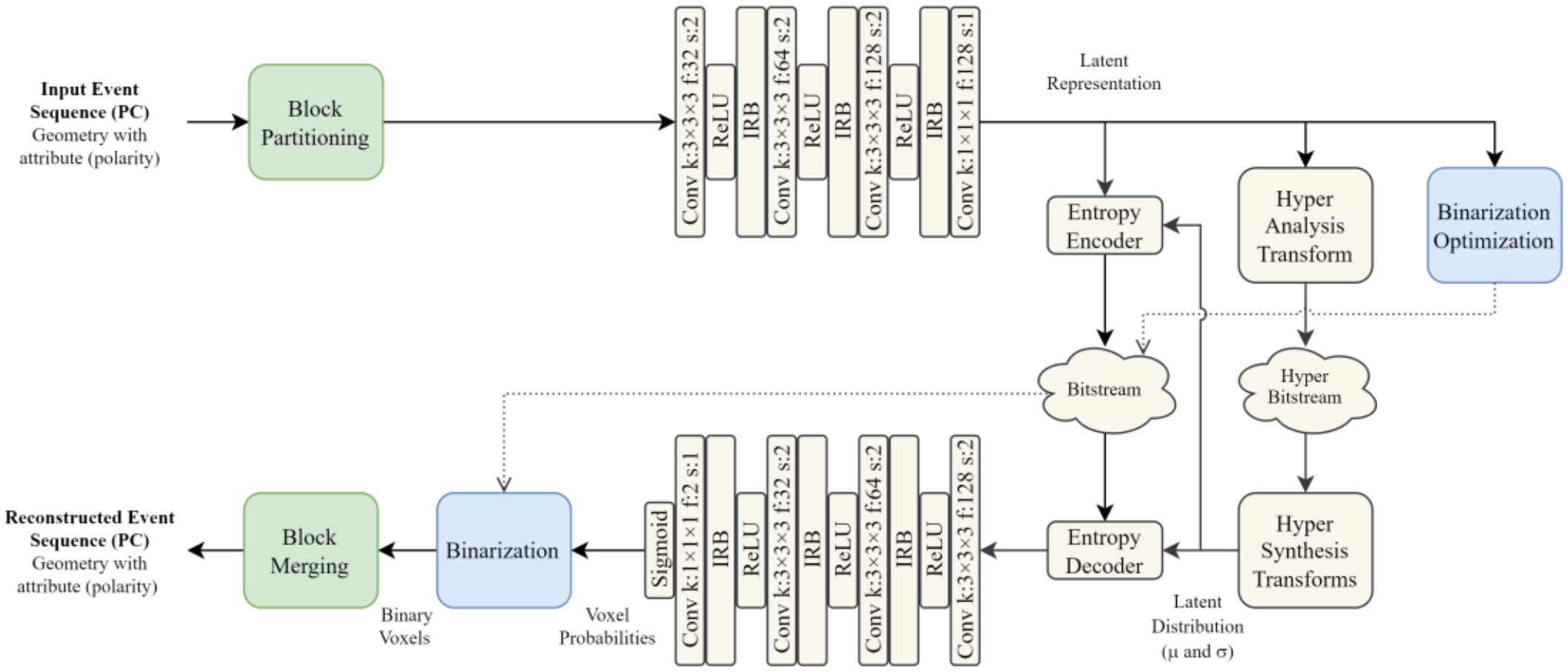

Fig. 2. DL-JEC coding model and binarization. The solid arrows refer to data moving along the various modules while the dotted arrows refer to coding parameters being inserted and extracted from the bitstream.

### A. DL-JEC Coding Model

As described in the previous section, DL-JEC processes event sequences segmented into blocks which are encoded using a single PC data structure. The DL-JEC coding model was designed by evolving the JPEG PCC coding model to make it jointly process the polarity information in addition to the geometry information as described below. By limiting the architectural modifications, DL-JEC offers a good level of compatibility with the JPEG PCC coding model, what may be critical for future standardization developments. The main DL-JEC coding model novelties are:

- **Model Input** – The DL-JEC coding model, represented in Fig. 2, adopts a sparse tensor representation [39], where the 3D block of voxels is represented by the set of coordinates corresponding to the non-empty voxels, and the set of corresponding features. Since the input voxels are binary-valued, the features of the corresponding sparse tensor are simply a single-channel vector of 'ones'. However, in the DL-JEC coding model, also the polarity information needs to be considered. Since the polarity information is also binary-valued, and it only exists for occupied voxels, the input sparse tensor is extended with a second features channel (also binary), where the POS polarity voxels are represented with 'ones' and NEG polarity voxels are represented with 'zeros'.
- **Latent representation** – After the first encoder layer, the geometry and polarity information are fused and represented jointly as a set of latent features. Throughout the remaining of the coding model, and up until the very last decoder layer, it is not possible to distinguish between features corresponding to the geometry and polarity information included in the bitstream. This means that DL-JEC, unlike JPEG PCC, does not offer component scalability, meaning that it is not possible to represent independently the geometry or the polarity information. However, since this scalability requirement is not relevant for event data, the novel joint processing allows the DL coding model to exploit additional relevant redundancies between the two polarities, thus contributing to improve the RD performance. The latent representation is entropy coded using an estimate of the mean (μ) and variance (σ) of the latents' distribution, which are determined by a variational hyperprior model, composed by a Hyper Analysis model and a Hyper Synthesis model, as described in [15].
- **Model Output** – Naturally, since the input consists of two channels, the DL decoding model, presented in Fig. 2, must also output a sparse tensor with two channels, thus requiring changing the number of filters of the last decoder layer from one to two. The first output channel will thus contain the occupancy probability of each voxel, while the second channel will contain the probability of the voxel having a POS polarity. Since the input data is binary, the decoding may be seen as a binary classification problem for each channel of each voxel. Given the output probabilities, the actual choice of whether the voxel is occupied or not and its corresponding polarity is made using a binarization strategy, as described in Section V.C.

Beside the coding model, the DL-JEC loss function has to be novel to take into account both the geometry and polarity information and train the model more effectively.

### B. Loss Function and Training

To ensure high compression efficiency, the novel DL coding model used by DL-JEC, including the coding and decoding models as well as the Hyper Analysis and Synthesis transforms presented in Fig. 2, is trained using a loss function that balances the total distortion ($D_{Total}$) and the estimated coding rate ($R$) through a Lagrangian multiplier, $\lambda$, as:

$$Loss\ Function = D_{Total} + \lambda \times R. \quad (1)$$

A separate DL coding model was trained for each RD point by varying $\lambda$. Five models were trained with $\lambda$ values of 0.00125, 0.0025, 0.005, 0.01, and 0.02, to cover a wide range of rates.

Since the DL-JEC model jointly codes geometry and polarity, a novel distortion metric was defined to take into consideration both the geometry and polarity distortions:

$$D_{Total} = (1-\omega)\cdot D_{Geometry} + \omega \cdot D_{Polarity}, \quad (2)$$

where $D_{Geometry}$ and $D_{Polarity}$ are the distortions computed for the spatiotemporal and polarity data, respectively, and $\omega$ is a weighting factor with a value set to 0.5. $D_{Geometry}$ is computed as the average binary classification error for each voxel defined by the Focal Loss (FL) [40] as follows:

$$FL(v,u) = \begin{cases} -\alpha(1-v)^{\gamma} \log v\,, & u = 1 \\ -(1-\alpha)v^{\gamma} \log(1-v)\,, & u = 0 \end{cases}, \quad (3)$$

where $u$ is the original voxel binary value, $v$ is the corresponding reconstructed voxel probability, and $\alpha = 0.5$ and $\gamma = 2$ are balancing weights. As for $D_{Polarity}$, since the polarity information is also binary, Focal Loss is used as well. However, due to the lossy nature of the DL-JEC coding model, some voxels may be reconstructed at locations that were originally empty. As there is no collocated ground truth polarity for such positions, the polarity of the nearest neighbor is used instead as the ground truth.

The training process follows a sequential approach, starting with the smallest value of λ (corresponding to the highest rate). The first model is trained with random initialization, while subsequent models use the weights from the previous model, reducing the training time and enhancing the RD performance. Early stopping is implemented to prevent overfitting, with patience of 25 epochs, halting training if validation loss does not improve.

The DL-JEC models were trained using the N-Caltech101 event dataset training and validation splits [19]. Each event sequence/object is scaled using TSF=128, which has been shown to maximize the compression performance for this dataset [10]. The different polarity duplicates are processed using a NN approach, meaning the polarity is determined by the neighborhood.

To limit the training computational complexity, the PCs used to represent the event data are divided into 64×64×64 blocks. To mitigate the negative impact of increased class imbalance and improve memory efficiency during training, blocks with fewer than 500 filled voxels or larger than 20,000 filled voxels were excluded, resulting in 34,942 blocks for training and 21,316 blocks for validation.

### C. *Binarization Optimization*

The DL-JEC decoding model outputs a set of voxels probabilities (see Fig. 2) in the range [0, 1] for the two channels: the first set represents the probability of a voxel being occupied while the second set represents the probability of the polarity being POS. These output probabilities require binarization to determine the final binary occupancy and polarity of the reconstructed points/events.

For the polarity channel, a fixed threshold of 0.5 is used, assigning POS polarity for probabilities of 0.5 or larger, and NEG polarity otherwise. However, for geometry, the use of a simple threshold is inefficient due to block density variations and reconstruction artifacts. Instead, an optimized Top-k binarization method is employed, where the $k$ voxels with the highest probabilities are filled as points/events at the decoder.

In JPEG PCC, the value of $k$ for regular PC coding is determined as: $k = N_{input} \times \beta$, where $N_{input}$ is the number of input points/events in the original block, and $\beta$ is optimized at the encoder for the best reconstruction quality, e.g., using the PSNR D1 geometry quality metric or an event data quality metric when coding event data. The $k$ value for each block is selected using some binarization optimization strategy and inserted in the bitstream for usage at decoding time. This process has a major impact on the final performance.

Binarization optimization at both JPEG PCC and DL-JEC encoders may be performed at two levels, block-level and sequence-level:

- **Block-level** – At block-level, the optimal value of $k$ is selected independently for each individual encoded block;
- **Sequence-level** – At sequence level, the optimal value of $k$ for each block is selected after merging the voxel reconstruction probabilities for all blocks of the sequence. In this approach, a sequence-level $k$ value is first determined based on a fidelity/quality metric or a computer vision performance metric. After establishing this sequence-level optimization, the optimum value of $k$ for each block is determined as the number of occupied voxels from the total reconstructed voxels in the sequence.

In this paper, the sequence-level approach is adopted for DL-JEC since a single PC is used. For JPEG PCC, each PC corresponds to a single polarity (two polarity streams exist) and thus a polarity-level approach is adopted. Sequence-level binarization optimization may allow reducing the negative impact of blocking artifacts; furthermore, it allows adopting some binarization strategies which are not suitable for block-level binarization optimization, particularly computer vision tasks; for example, it does not make sense to classify an event by individually classifying its multiple blocks.

As the number of reconstructed points/events has a major impact on both the compression efficiency and classification performance, this paper proposes three binarization optimization strategies for defining the value of $k$ as follows:

- **Count-optimized binarization (CoB)** – The value of $k$ is defined as $k = N_{input}$ where $N_{input}$ is the number of input points/events in the original block. This binarization strategy ensures that the decoder reconstructs event data with the same number of events as the input, which is particularly beneficial for tasks sensitive to event count. A more restrictive option of this binarization strategy can be used to accurately reconstruct both the total event count as well as the split over the two event polarities by considering the number of both POS events ($N_{POS}$) and NEG events ($N_{NEG}$) as extra information in the bitstream.
- **Quality-optimized binarization (QuB)** – The value of $k$ is optimized at the encoder by maximizing the

reconstruction quality, according to an appropriate quality metric such as the PSNR D1 geometry quality metric [35][41] for PCC, or the proposed joint geometry and polarity PSNR E2E metric (proposed in the next section) for event data coding. This optimization may be performed at block, polarity or sequence-level. In this paper, the PSNR E2E quality metric has been used for both DL-JEC and JPEG PCC QuB binarization optimization at sequence/polarity-level. It is worthwhile to mention that QuB optimization does not explicitly consider any downstream computer vision tasks performance, it is just a data fidelity optimization.

- **Classification-optimized binarization (ClB)** – The value of $k$ is optimized at the encoder by maximizing the resulting classification accuracy according to selected metrics, notably the Top-1 and Top-5 accuracy metrics used in this paper. The optimization process prioritizes Top-1 accuracy, by selecting a value of $k$ that ensures that the event sequence classification is correct (the right class is predicted); if such a value of $k$ cannot be found for Top-1, the binarization strategy continues by choosing the $k$ value ensuring that the correct class is one of the Top-5 classes. If neither condition may be satisfied, the value of $k$ corresponding to the maximum probability assigned to the correct class is selected. Since classification has to be performed using the full event sequence (and not a single polarity), this strategy can only be used for DL-JEC and not JPEG PCC since the coding process is performed independently for the two polarities. Naturally, optimization binarization strategies for other computer vision tasks may be developed using the appropriate performance metrics.

The novel ClB and CoB strategies are designed specifically for event data classification and tasks affected by the number of events, respectively. It is important to note that these strategies do not require any additional training or modifications to the DL coding model, thus ensuring effective integration without modifying the decoder.

## VI. Experimental Setup

This section describes the experimental setup for assessing the performance of the proposed DL-JEC solutions for coding event data as a single PC in comparison to conventional and DL-based PCC codecs for coding event data as two separate PCs. For a more complete assessment, the experimental setup considers three event data classification pipelines as follows:

- **Original Event Data Classification pipeline** – Classifies the original floating-point precision event data (*x, y, t, p*). This pipeline serves as the reference for the classification performance since no artifacts are introduced in the events input to the classifier.
- **Voxelized Event Data Classification pipeline** – Converts event data from floating-point precision to integer precision. The goal is to assess the impact of lossy voxelization on event data classification. Since this pipeline does not involve any coding, no compression artifacts are present.
- **Decompressed Event Data Classification pipeline** – Classifies reconstructed events after lossless (post-voxelization) or lossy PC encoding/decoding solutions, notably:
  1. **Proposed lossy DL-JEC** solution coding event data as a single PC.
  2. **Lossless and lossy (Octree) G-PCC** (TMC13 reference software, version v21, https://github.com/MPEGGroup/mpeg-pcc-tmc13) coding event data as two separate PCs.
  3. **Lossy JPEG PCC** (software version 4.1) coding event data as two separate PCs.

The EST classifier is used for all pipelines, including both the proposed coding solution as well as the conventional and DL-based benchmarks, without any specific retraining (other than the training reported in Section III with the training and validation splits of the N-Caltech101 original event data).

The N-Caltech101 event dataset test split [19] was chosen in this paper for performance assessment, since it is a commonly used dataset for event data classification. The test split of the N-Caltech101 event dataset, composed of 1,741 event sequences organized in 101 classes, is used in the experiments for all the considered event data classification pipelines, i.e., original, voxelized, and decompressed domains; the spatial resolution is 256×256 and the temporal duration 300 ms. In addition, the IBM DVS Gesture event dataset [20] is used to assess the DL-JEC generalization capability to a different event dataset and computer vision task; the spatial resolution is 128×128 and the (average) temporal duration 6 seconds.

The performance metrics for assessing the event data compression (RD) and classification performances are:

- **Rate** – Corresponds to the total coding rate obtained when compressing the input event sequence data, measured in bits-per-(original) event (bpe); for the two PCs coding solutions, the rate is the sum of the rate for the two polarity streams.
- **Peak Signal-to-Noise Ratio Event-to-Event Distance (PSNR E2E)** – Since there is no quality metric available in the literature specifically developed for lossy event data coding, a novel quality metric is proposed here. The new metric is inspired by and generalizes the PSNR D1 metric, adopted in both the MPEG Common Test Conditions (CTC) [41] and JPEG CTTC [35] for PC geometry quality. Assuming $A$ and $B$ represent the reference and reconstructed event sequences/objects, respectively, the proposed PSNR E2E metric is computed as:
  1. Compute $MSE_{A,B}$ and $MSE_{B,A}$:

$$MSE_{A,B} = \frac{e^{POS}_{A,B} + e^{NEG}_{A,B}}{|A|}, \text{ and } MSE_{B,A} = \frac{e^{POS}_{B,A} + e^{NEG}_{B,A}}{|B|} \quad (3),$$

where $e^{POS}_{A,B}$ ($e^{NEG}_{A,B}$) is the sum of the squared Euclidean distances between each POS (NEG) event in $A$ and its POS (NEG) nearest neighboring event in $B$; the consideration of the polarities is a key difference to PSNR D1. Similarly, $e^{POS}_{B,A}$ and $e^{NEG}_{B,A}$ are computed considering events in $B$ and their nearest neighbors in A.

$|A|$ and $|B|$ represent the total number of events in $A$ and $B$, respectively. It is important to notice that the Mean Squared Error (MSE) value defined here becomes equivalent to the one used for PSNR D1 when only one polarity is considered.

2. Compute PSNR E2E:

$$PSNR\ E2E_{seg} = 10log\frac{Px^2+Py^2+Pt^2}{max(MSE_{A,B},MSE_{B,A})} \quad (4),$$

where $Px$, $Py$ and $Pt$ are related to the maximum resolution used to represent the spatial and temporal coordinates and correspond to the next power of two of the maximum value for each of the $(x, y, t)$ event data resolutions (total number of bins), respectively. Unlike for PC, where the resolution for the $(x, y, z)$ coordinates is typically uniform, for event data, the temporal precision may be much larger than the spatial precisions, meaning that the distortion along the spatial and temporal dimensions would be considered in a different way. For this reason, it is proposed to compute PSNR E2E using as reference a version of the original event which is temporally scaled using an appropriate TSF parameter, to approximate the temporal and spatial precisions and properly balance the impact of the distances/MSE in the spatial and temporal domains. Moreover, the original event sequence is used without any duplicates, meaning Same Polarity and Different Polarity Duplicates Removal modules are applied to define the event data ground truth/reference. Naturally, the same TSF scaling is applied to the reconstructed event sequence which is already duplicates free after the PC to event conversion process. Therefore, the TSF value is chosen such that the temporal resolution computed as the event sequence duration multiplied by TSF is similar to the largest spatial resolution:

- For the N-Caltech101 dataset, with a spatial resolution of 256×256 and a temporal duration of 300 ms, TSF comes to 512, resulting in $Px = Py = Pt$ =256.
- For the IBM DVS Gesture dataset, with a spatial resolution of 128×128 and an (average) temporal duration of 6 s, TSF comes to 16, resulting in $Px = Py = Pt$ =128.

If the event sequence is too long to be assessed in a single segment, the final PSNR E2E metric may be computed as the average of PSNR E2E computed over a set of temporal segments with a selected duration, $t_{seg}$, using a TSF value such that the segment temporal resolution, $R_{seg}$, matches the spatial resolution. The computation of PSNR E2E by segments may also be useful when the original and distorted event sequences are available in a progressive way. In these situations, the duration of each event segment may be determined by the frequency in which a value of the PSNR E2E metric is required.

- **Top-k Classification Metric** – Measures the percentage of test examples where the ground truth label is among the k predicted classification labels with the highest probabilities. For this paper, k is set to 1 (Top-1) and 5 (Top-5).

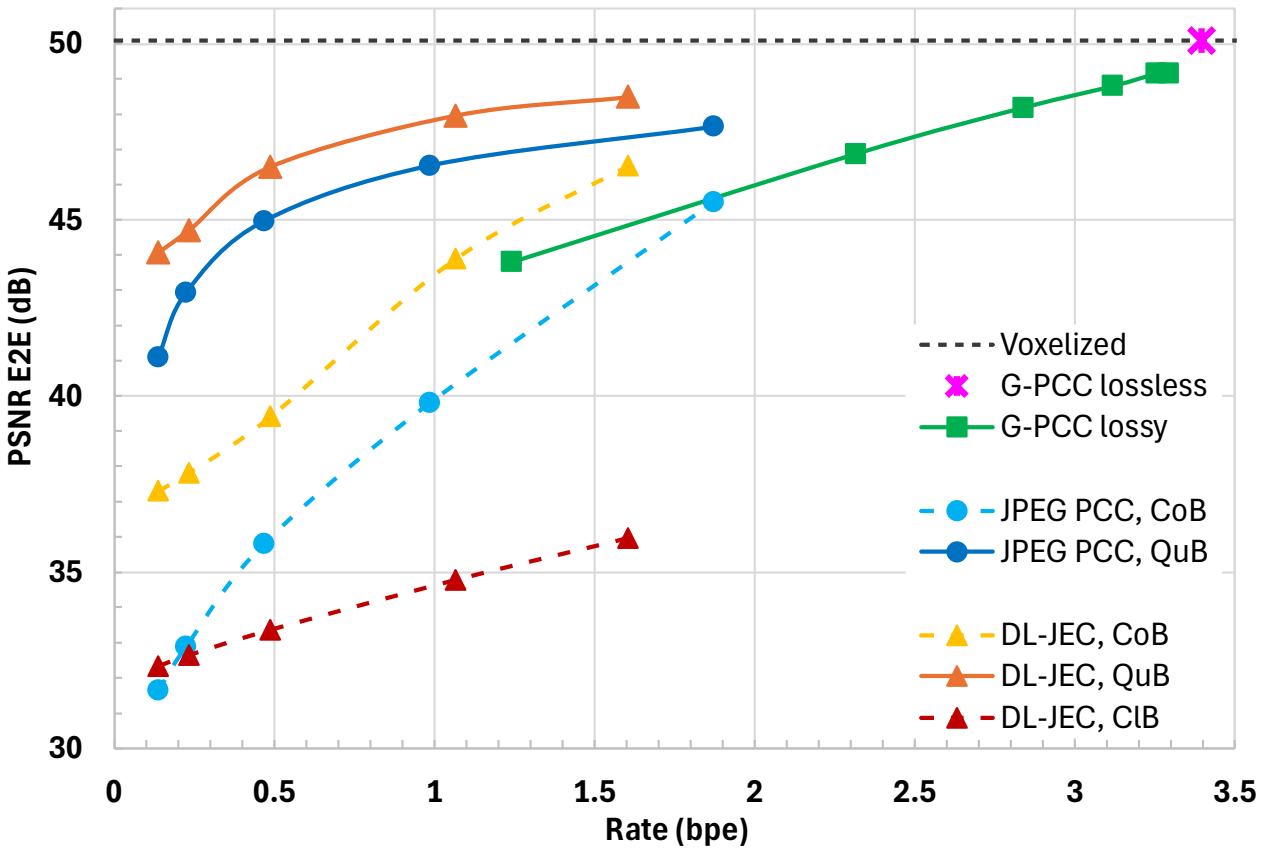

Fig. 3. RD performance assessment using PSNR E2E as event quality metric for the N-Caltech101 dataset test split.

The next section will report and discuss the performance assessment obtained under this experimental setup.

## VII. Performance Assessment

This section reports and discusses the DL-JEC performance, using all reference classification pipelines, including original and voxelized event data. First, the DL-JEC event data compression performance is analyzed for all binarization strategies and compared with both conventional and DL-based coding benchmarks. After, a similar assessment is made for the classification performance using the EST classifier on the reconstructed event data.

### A. *Event Data Compression Performance*

This subsection presents and discusses the event data compression performance. DL-JEC is compared with state-of-the-art solutions that use two separate PCs and the behavior of the novel binarization strategies is analyzed. The compression results for the N-Caltech101 event dataset test split are presented in Fig. 3, where PSNR E2E values are plotted as a function of the rate. Both JPEG PCC and DL-JEC use the same coding configurations, notably a block size of 256 and a scaling factor of 1. A block size of 256 enables each input N-Caltech101 event sequence to fit in a single coding block.

#### 1) *DL-JEC with Different Binarization Strategies*

The DL-JEC compression performance is analyzed for the various binarization strategies. The PSNR E2E for voxelized events (without coding) is shown in Fig. 3 as a horizontal line, and represents the reconstruction quality achieved solely by the lossy voxelization process (i.e., without compression artifacts). The results in Fig. 3 show:

- DL-JEC QuB RD performance considerably outperforms both DL-JEC CoB and ClB RD performances for the PSNR E2E quality metric. This is expected since QuB optimization targets the reconstruction quality, unlike CoB and ClB, which are optimized for event-count and classification performance, respectively.
- DL-JEC CoB offers better RD performance than DL-JEC ClB, since the PSNR E2E metric is influenced by the

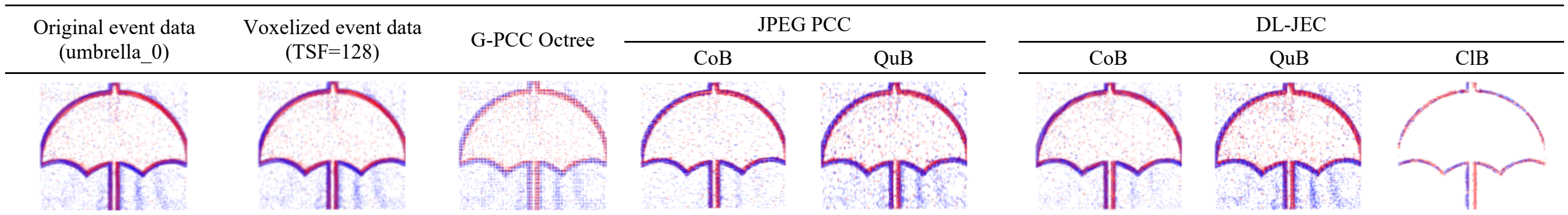

Fig. 4. Original, voxelized, and reconstructed event data sequence from the N-Caltech101 dataset, using three codecs: G-PCC lossy (with second lowest rate), and the highest rates from JPEG PCC (CoB and QuB), and DL-JEC (CoB, QuB, and ClB). Similar rates were selected to ensure a fair comparison. Red and blue indicate NEG and POS events, respectively.

number of events. It was observed that ClB reconstructs event data with fewer points/events than CoB and QuB, as shown in Fig. 4.

As expected, these findings demonstrate that DL-JEC compression performance depends on the chosen binarization strategy and a quality friendly binarization optimization strategy boosts the compression performance.

*2) DL-JEC versus Two PC-based Benchmark Codecs*

The proposed lossy DL-JEC codec (coding a single PC) is compared with the conventional G-PCC (lossless and lossy) and the DL-based JPEG PCC codec, applied to event data represented as two separate PCs, one per polarity. As a reference, Fig. 3 shows the G-PCC lossless rate as a pink cross since lossy compression above this rate is meaningless; for lossless coding, the PSNR E2E value is the same as for the voxelized pipeline since a losslessly coded voxelized version of the event data is compared with the original (non-voxelized) event data taken as reference.

The results in Fig. 3 show:

- DL-JEC QuB offers high quality for rates much below the ones required for G-PCC lossless coding.
- DL-JEC QuB achieves significant RD performance gains over G-PCC lossy for similar rates, though G-PCC can reach higher qualities at much higher rates.
- DL-JEC offers consistently better RD performance than JPEG PCC, for all rates and the same binarization strategies (QuB and CoB).
- The RD performance gap between DL-JEC CoB and JPEG PCC CoB is larger than between DL-JEC QuB and JPEG PCC QuB, due to JPEG PCC generating duplicate events with different polarities (which are removed).

Overall, the compression results in Fig. 3 clearly show that the proposed DL-JEC solution is better than G-PCC lossy and JPEG PCC, notably when optimizing the binarization for a reconstruction fidelity metric, i.e., QuB for PSNR E2E.

*3) Computational Complexity Comparison*

The average inference times (including encoding and decoding) were compared for the highest-rate coding models of both DL-JEC and JPEG PCC, evaluated on the N-Caltech101 event dataset test split. The experiments were conducted on a workstation equipped with an Intel Core i7-14700k CPU @ 5.60 GHz, an NVIDIA GeForce RTX 4090 (24 GB) GPU, and 64 GB of RAM, running Debian 12. The results show that DL-JEC achieves lower average encoding and decoding times (4.87 and 1.29 seconds, respectively) compared to JPEG PCC (6.85 and 1.62 seconds), thus demonstrating its better computational performance.

### *B. Decompressed Domain Event Data Classification Performance*

This subsection presents and discusses the Top-1 and Top-5 classification performances for various solutions. The classification results for the N-Caltech101 event dataset test split are presented in Fig. 5. The proposed DL-JEC codec is compared with the benchmarks, notably for the three event classification pipelines described in Section VI. The classification results for the original and voxelized classification pipelines are represented by horizontal lines in Fig. 5; the proximity of the two lines suggests that voxelization does not significantly impact the classification performance.

*1) DL-JEC with Different Binarization Strategies*

When comparing DL-JEC classification performance for the three binarization strategies, the results in Fig. 5 show:

- DL-JEC QuB has lower classification performance than DL-JEC CoB and DL-JEC ClB, because it is optimized for reconstruction quality rather than classification performance.
- DL-JEC QuB and DL-JEC CoB can achieve a classification performance close to the original and voxelized classification performances, notably at higher rates; however, DL-JEC CoB outperforms DL-JEC QuB at lower and medium rates.
- DL-JEC ClB provides the best classification performance, surpassing both the original and voxelized (same as G-PCC lossless) classification performance at all rates for Top-1, and matching it for Top-5. This unexpected behavior is explained by the fact that the N-Caltech101 event dataset test split is rather noisy and the ClB binarization acts like a denoising filter, offering the EST classifier ‘cleaner’ objects, as shown in Fig. 4.

As expected, these findings demonstrate that DL-JEC classification performance depends on the selected binarization strategy and a classification friendly binarization strategy boosts the classification performance.

*2) DL-JEC Versus Benchmark Two PC-based Codecs*

The proposed lossy DL-JEC codec (coding a single PC) is compared with the conventional G-PCC (lossy and lossless) and the DL-based JPEG PCC codec, using two separate PCs, in terms of decompressed domain classification performance. The results in Fig. 5 show:

- DL-JEC offers a classification performance close or even

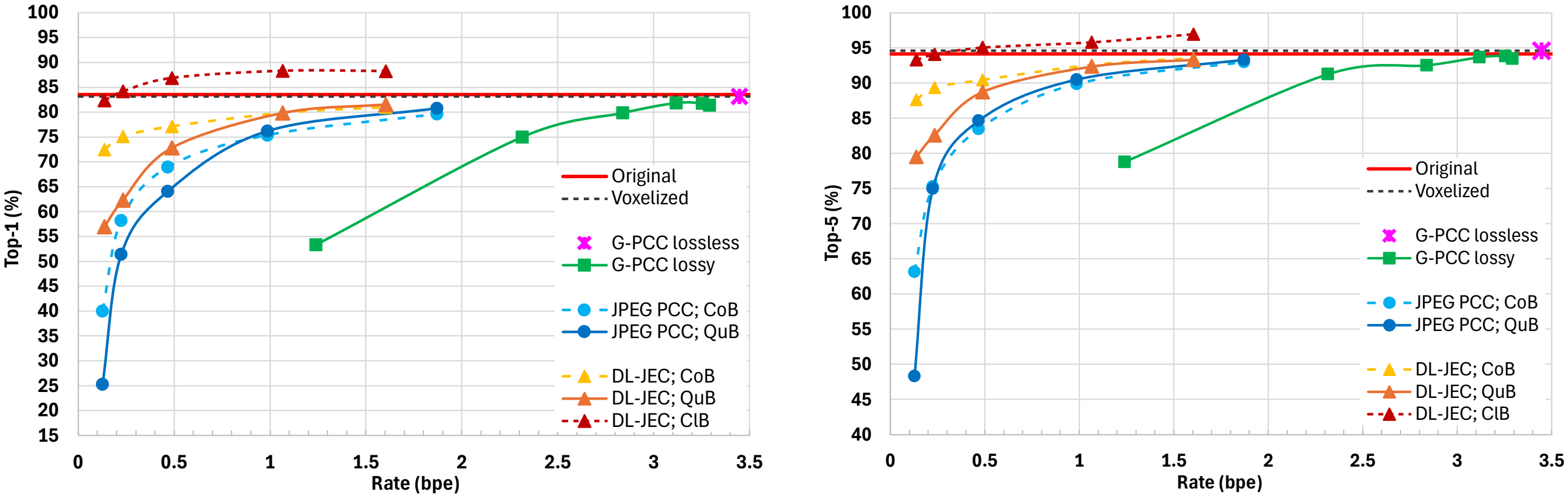

Fig. 5. Classification performance for the N-Caltech101 dataset test split using Top-1 (left) and Top-5 (right) accuracy metrics for the EST event data classifier and all classification pipelines.

better than the original event data classification, with a much lower rate than G-PCC lossless.

- DL-JEC consistently achieves significant classification performance gains (for both Top-1 and Top-5) over G-PCC lossy. The worse DL-JEC ClB classification performance is comparable to the best G-PCC lossy classification performance.
- DL-JEC consistently achieves significant classification performance gains (for both Top-1 and Top-5) over JPEG PCC, for all binarization strategies. The worse DL-JEC ClB classification performance is comparable to the best JPEG PCC classification performance for both CoB and QuB.
- JPEG PCC clearly outperforms G-PCC lossy in terms of the decompressed domain classification performance, for both Top-1 and Top-5, particularly at low to medium rates.
- Similar to DL-JEC, JPEG PCC CoB offers better classification performance than JPEG PCC QuB since the test and binarization optimization metrics match better.

While a binarization strategy like ClB may lower the PSNR E2E quality (due to metrics mismatch), it significantly improves the classification performance, especially at low data rates. This highlights the importance of selecting the right binarization optimization strategy depending on the specific event data task/target: either maximizing the machine-oriented task performance, e.g., for classification like ClB in this paper, or maximizing the reconstruction quality, using QuB.

The reported results demonstrate that the use of joint coding solutions for event data, where spatiotemporal and polarity information are coded together (instead of in a component scalable manner), increase both the event data compression performance and the decompressed domain event data classification performance by better exploiting the correlation between all event data components. Furthermore, although current research activity on state-of-the-art event data coding primarily focuses on lossless solutions, these results demonstrate that the usage of lossy compression can achieve comparable performance for tasks like event data classification, with significant rate reductions. This is a major achievement which paves the way to lossy event data coding, notably in the context of the JPEG XE standardization project.

### C. *Generalization Capabilities*

This subsection reports and analyses the compression and classification performances for an additional dataset, the IBM DVS Gesture event dataset [20], with the goal to assess the generalization capabilities of the proposed DL-JEC solution and novel binarization strategies targeting reconstruction and classification tasks. To assess generalization, the IBM DVS Gesture test split is coded with the DL-JEC coding models still trained with the N-Caltech101 training and validation splits.

The IBM DVS Gesture dataset is acquired using a static event-based camera, with a spatial resolution of 128×128 pixels, which captures event data from 29 participants performing hand and arm gestures under three different lighting environments, notably natural daylight, fluorescent lighting, and LED lighting. The dataset includes 1,342 event sequences, with an average duration of 6 seconds, organized into 11 different types of gestures (classes), with 122 event sequences per gesture. This event dataset is organized into training and test splits, including 1,078 sequences from 26 people, and 264 sequences from the remaining 3 people, for the training and test splits, respectively. Each event sequence is classified into one of 11 gesture classes. IBM DVS Gesture is one of the most used datasets for developing and testing event-based methods for action/gesture recognition.

The use of the IBM DVS Gesture event dataset with the G-PCC (lossless and lossy), JPEG PCC, DL-JEC codecs and the EST classifier requires the following configurations:

- Since the IBM DVS Gesture dataset only defines training and test splits, the training set of 1,078 sequences was further divided into training (80%) and validation (20%) splits, by randomly selecting sequences from each gesture class in this proportion. This division ensures consistent class representation in both splits, allowing for reliable EST classifier model training.
- For coding and decoding event data as single PC using DL-JEC, each event sequence in the IBM DVS Gesture test split was converted into a single PC using the same TSF=128 as used for the N-Caltech101 test split. After conversion, DL-JEC was then applied using the same five models jointly trained with the geometry and polarity of the N-Caltech101 training and validation splits, to assess the DL-JEC generalization capabilities.

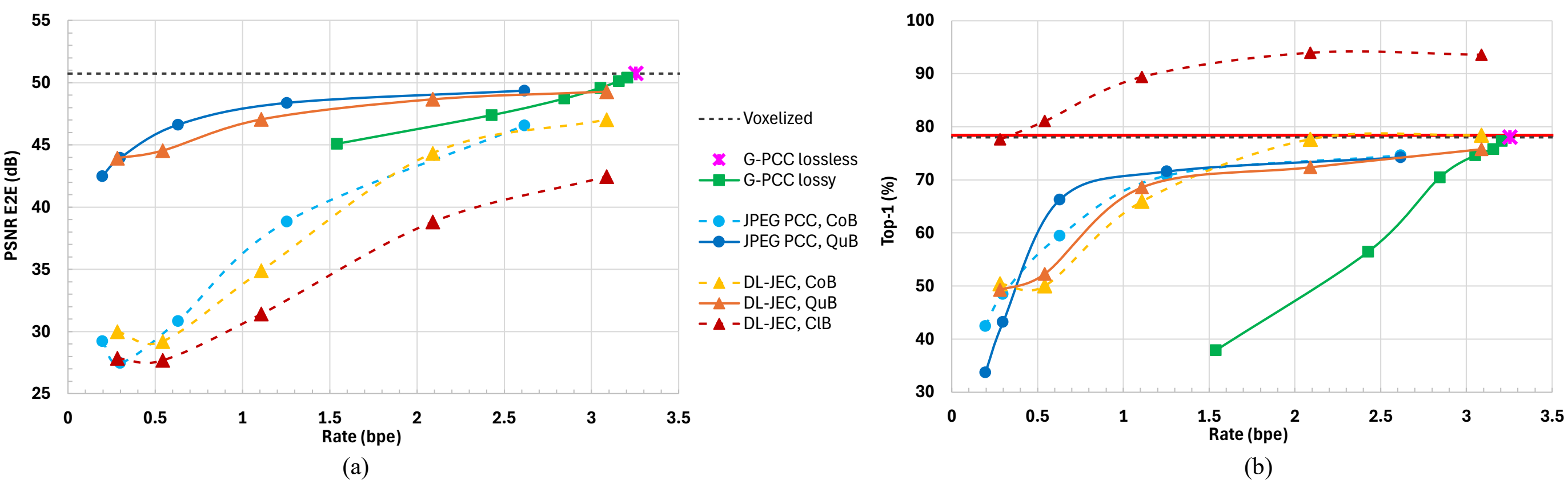

Fig. 6. Compression and classification performances for the IBM DVS Gesture dataset split: (a) PSNR E2E; and (b) Top-1 classification accuracy for the EST event data classifier and all classification pipelines.

- For coding event data as two separate PCs using G-PCC (lossless and lossy) and JPEG PCC, each event sequence from the IBM DVS Gesture test split was converted into two PCs (one for each polarity) with TSF=128 as previously for N-Caltech101 test split.
- Since the IBM DVS Gesture event dataset has different classes from the N-Caltech101 event dataset, the EST classifier was, naturally, trained using the IBM DVS Gesture training and validation splits and used for testing all classification pipelines with the IBM DVS Gesture test split.

Fig. 6a shows the IBM DVS Gesture test split PSNR E2E compression performance for the proposed DL-JEC using the three proposed binarization strategies, compared with the selected coding benchmarks. The same coding configurations to code the N-Caltech101 test split with JPEG PCC and DL-JEC, notably a block size of 256 and a scaling factor of 1, are now used to code the IBM DVS Gesture dataset test split. Contrary to N-Caltech101 where each event sequence fits in a single block, for IBM DVS Gesture, each event sequence has to be coded as multiple blocks due to its longer time duration. For the QuB binarization strategy, both DL-JEC and JPEG PCC were binarized to maximize PSNR E2E at sequence and polarity levels, respectively. The PSNR E2E compression results allow the following observations:

- Overall, DL-JEC offers good RD performance for the IBM DVS Gesture test split, even if coded with models trained with the N-Caltech101 training split, thus showing the DL-JEC good generalization capabilities.
- DL-JEC QuB outperforms G-PCC lossy in PSNR E2E performance. Also, DL-JEC attains high reconstruction quality for bitrates significantly lower than the G-PCC lossless rate.
- DL-JEC QuB achieves a comparable PSNR E2E performance regarding JPEG PCC. The explanation for the slightly better JPEG PCC RD performance in relation to the DL-JEC performance for the IBM DVS Gesture test split (which was not observed for the N-Caltech 101 test split), is related to the relative characteristics of the training datasets (N-Caltech101 for DL-JEC and JPEG PCC training set for JPEG PCC) and the IBM DVS Gesture test split. Table 1 demonstrates the differences between the datasets by characterizing them in terms of the Density Factor (DF) [42] used by MPEG, which measures how dense are the datasets by considering the proximity of neighboring points. Table 1 clearly shows that the IBM DVS Gesture test split density characteristics are closer to the JPEG PCC training dataset (which codes each polarity independently and not all events together) than to the N-Caltech101 training dataset characteristics. This indicates that DL-JEC has been trained on more dense data than encountered during testing with IBM DVS Gesture in comparison with JPEG PCC. As a result, the DL-JEC models are less 'fitted' to the IBM DVS Gesture test split than the JPEG PCC models, thus justifying the relative performance. Nevertheless, DL-JEC offers almost equivalent compression performance to JPEG PCC, showing good generalization capabilities.

TABLE 1
EVENT DATASETS DENSITY.

| | JPEG PCC training split | N-Caltech101 training split | IBM DVS Gesture | | |
|---|---|---|---|---|---|
| | | | Test split | Test split (POS) | Test split (NEG) |
| Mean | 0.45 | 0.66 | 0.32 | 0.17 | 0.19 |
| Standard deviation | 0.22 | 0.24 | 0.23 | 0.09 | 0.14 |

Fig. 6b shows the Top-1 classification performance for DL-JEC and the selected coding benchmarks for the IBM DVS Gesture dataset test split which allow the following observations:

- Although the DL-JEC coding model was trained on the N-Caltech101 training and validation splits, the overall classification performance behavior for the tested coding solutions for the IBM DVS Gesture dataset is similar to the behavior previously obtained for the N-Caltech101 dataset.

| Original event data (User29_natural) | Voxelized event data (TSF=128) | G-PCC Octree | JPEG PCC | | DL-JEC | | |
|---|---|---|---|---|---|---|---|
| | | | CoB | QuB | CoB | QuB | ClB |

Fig. 7. Original, voxelized, and reconstructed event data sequence from the IBM DVS Gesture dataset, using three codecs: G-PCC Octree (with fourth lowest rate), and the highest rates from JPEG PCC (CoB and QuB), and DL-JEC (CoB, QuB, and ClB). Similar rates were selected to ensure a fair comparison. Red and blue indicate NEG and POS events, respectively.

- Both DL-JEC and JPEG PCC are able to achieve Top-1 accuracy close to the original and voxelized classification pipelines, at rates which are well below the G-PCC lossless rate.
- DL-JEC with the ClB binarization strategy delivers the highest Top-1 accuracy across all bitrates, clearly surpassing the original and voxelized pipelines, even for rates eight times smaller than the G-PCC lossless rate. As for the N-Caltech101 dataset, this can be explained by the noise reduction effect obtained with the ClB binarization which allows the classifier to better focus on the gesture part of the event sequence; this effect may be observed in Fig. 7 for a IBM DVS Gesture event sequence example, as previously observed in Fig. 4 for a N-Caltech101 event sequence example.

Unlike the N-Caltech101 dataset, which exhibits three saccadic movements, the IBM DVS Gesture dataset features varied temporal dynamics due to the different types of hand and arm movements across gesture classes, making the two datasets rather different. Although the compression and classification performance behaviors are clearly consistent for the two studied datasets, showing that the proposed methods generalize well, the performance may be further improved by specifically optimizing the TSF parameter for different event datasets.

These findings underscore the importance of training future DL-JEC models on more diverse event-based datasets if models able to generalize to datasets with different characteristics are targeted.

The average inference times (encoding and decoding) for the IBM DVS Gesture test split dataset were compared using JPEG PCC and DL-JEC under identical computational settings: Intel Core i7-14700K CPU @ 5.60 GHz, NVIDIA RTX 4090 (24 GB) GPU, 64 GB RAM, running Debian 12. For part of the IBM DVS Gesture dataset, DL-JEC has lower average encoding and decoding times (7.39 and 2.21 seconds, respectively) when compared with JPEG PCC (12.43 and 3.08 seconds, respectively). These results consistently demonstrate DL-JEC's lower computational complexity compared with JPEG PCC across different datasets.

## VIII. Conclusion

This paper proposes a novel lossy DL-based solution for efficient event data coding, named DL-JEC. By integrating both spatiotemporal and polarity information into a single PC, DL-JEC is able to enhance the RD performance compared to conventional G-PCC and DL-based JPEG PCC used for coding event data as two separate PCs (one for each polarity). The innovative loss functions and binarization optimization strategies proposed in this work allow for tailored performance for various computer vision tasks, particularly classification. The experimental results demonstrate that DL-JEC not only achieves superior RD performance but also offers high computer vision task performance at much lower rates than lossless coding.

DL-JEC's unified representation opens the door for efficient compressed domain event data classification without compromising classification performance or other computer vision tasks performance. By directly feeding the single latent representation into a compressed domain classifier, the need for a full decoder is eliminated, significantly reducing the overall complexity and highlighting the potential of lossy DL-based coding techniques to advance the field of event-based vision systems.

Further future work may extend the proposed event data coding and processing framework to additional vision tasks, such as semantic segmentation, to assess the influence of compression on downstream performance. Given that the EST classifier used in this paper depends on a specific learning strategy for integrating event data into the classification network, future research will explore alternative event representations, notably to boost the method's generalization capabilities. In addition, DL-JEC may be trained on a diverse collection of event-based datasets with varying characteristics, particularly in terms of event density, to enhance its robustness and adaptability across different scenarios.